\documentclass[11pt,a4paper]{article}
\usepackage[hyperref]{emnlp2020}
\usepackage{times}
\usepackage{latexsym}
\usepackage{graphicx}
\usepackage{amsmath}
\usepackage{booktabs}
\usepackage{multirow}
\usepackage{url,paralist}
\usepackage{hyperref}

\usepackage{microtype}

\aclfinalcopy 

\title{What Do Position Embeddings Learn?\\ An Empirical Study of Pre-Trained Language Model Positional Encoding}

\author{Yu-An Wang \quad Yun-Nung Chen\\
  Department of Computer Science and Information Engineering \\
  National Taiwan University, Taipei, Taiwan \\
  \texttt{r08922019@csie.ntu.edu.tw} \quad
  \texttt{y.v.chen@ieee.org} \\}

\date{}

\begin{document}
\maketitle
\begin{abstract}
In recent years, pre-trained Transformers have dominated the majority of NLP benchmark tasks. 
Many variants of pre-trained Transformers have kept breaking out, and most focus on designing different pre-training objectives or variants of self-attention. 
Embedding the position information in the self-attention mechanism is also an indispensable factor in Transformers however is often discussed at will. 
Therefore, this paper carries out an empirical study on position embeddings of mainstream pre-trained Transformers, which mainly focuses on two questions:
1) Do position embeddings really learn the meaning of positions?
2) How do these different learned position embeddings affect Transformers for NLP tasks? 
This paper focuses on providing a new insight of pre-trained position embeddings through feature-level analysis and empirical experiments on most of iconic NLP tasks.
It is believed that our experimental results can guide the future work to choose the suitable positional encoding function for specific tasks given the application property.\footnote{The source code is available at: \url{https://github.com/MiuLab/PE-Study}}
\end{abstract}

\section{Introduction}
Word ordering often determines the meaning of a sentence; therefore how to utilize the position information of a word sequence has been an important topic in NLP and widely investigated recently.
A common approach for modeling word ordering is to use recurrent neural networks (RNN), such as long short-term memory (LSTM) \citep{hochreiter1997long} or gated 
recurrent unit (GRU) \citep{chung2014empirical}, which use a hidden state to represent the information of an ordered sequence and update model weights by backpropagation through time (BPTT) \citep{werbos1990backpropagation}; thus the ordering information can be modeled by this structure. 
However, RNN and BPTT are very inefficient in modern GPU computation due to the difficulty of parallelization with the time dependency.
To solve this problem, recent work, such as convolutional seq2seq \citep{gehring2017convolutional} and Transformers \citep{vaswani2017attention} which apply convolutional neural network (CNN) \citep{lecun1995convolutional} and self-attention respectively, succeed to eliminate the time dependency to take the computational advantage of GPU. 
Instead of storing the information of ordered sequences, these models utilize the position information by using a feature-level positional encoding. For example, convolutional seq2seq proposed learnable position embeddings to represent the positions in a sequence.

Recently, various pre-trained Transformer language models keep breaking state-of-the-art results in numerous NLP tasks. 
There are many different ways to pre-train a Transformer language model.
For example, using an encoder, decoder, or the whole part of the Transformer, adapting the self-attention masks, or training with different objectives \citep{devlin2018bert, liu2019roberta, radford2018improving, radford2019language, lewis2019bart, raffel2019exploring, yang2019xlnet}. 
However, in terms of positional encoding, most work only used a learned position embedding which is originally proposed in convolutional seq2seq \citep{gehring2017convolutional} without any analysis, even different objectives may learn completely different position information.

Motivated by the above observations, our goal is to investigate what position information the pre-trained Transformers could learn under different settings.
We conduct a deep analysis of the learned position embeddings among three iconic pre-trained Transformer language models: BERT \citep{devlin2018bert}, RoBERTa \citep{liu2019roberta} and GPT-2 \citep{radford2019language}.
To examine the performance of different NLP types, we conduct the experiments on text classification, language modeling, and machine translation, and empirically analyze and explain the meaning and influence of position embeddings from different aspects.

The contributions of this paper are 3-fold:
\begin{itemize}
\item This paper is among the first study that provides a complete analysis about what learned position embeddings capture in different pre-trained models.
\item This paper empirically examines the performance of different position embeddings for many NLP tasks.
\item This paper connects the empirical performance with the task property based on the analysis, providing the guidance of the future work for choosing the suitable positional encoding method in the target task.
\end{itemize}

\section{Related Work}
The concept of using position embedding on position-insensitive models was first proposed by convolutional seq2seq \citep{gehring2017convolutional}, which built an encoder-decoder architecture on convolutional neural networks.
\citet{vaswani2017attention} proposed Transformers that used the self-attention mechanism in the basic blocks. 
Because the attention mechanism is \emph{position-insensitive}, it proposed a pre-defined sinusoidal function as positional encoding. 
Pre-trained language models became a trend among many NLP tasks after \citep{peters2018deep} introduced ELMo. Affected by ELMo, OpenAI GPT \citep{radford2018improving} is the first pre-trained language model using a Transformer architecture, then many different variant of pre-trained Transformer including BERT \citep{devlin2018bert}, RoBERTa \citep{roberts-2005-learning} and GPT-2 \citep{radford2019language} started evolving the researches of NLP tremendously.
In Transformers, the attention values are the same in each input position.
Thus, \citet{shaw2018self} proposed a relative position representation in the attention level to address this issue.
\citet{dai2019transformer} used a segment-level recurrence mechanism on Transformers and also utilized an adaptive version of relative position embeddings inspired by \citet{shaw2018self}.
Furthermore, \citet{wang2019encoding} extended the embedding space from real numbers to complex values
, and also proposed a new learnable positional encoding function instead of a simple position embedding mapping.

\section{Transformer}
Transformer is an encoder-decoder sequence-to-sequence model proposed by \citet{vaswani2017attention}.
In the architecture, Transformer is composed of self-attention blocks that are position-insensitive modules. 
Therefore, a positional embedding should be considered together with the NLP tasks. 
To elaborate on the experiments we conduct, this section briefly introduces Transformers.

\paragraph{Input Representation}
Due to the property of position-insensitive in the attention module, the input representations should also contain the position information. 
In Transformers~\cite{vaswani2017attention}, a word embedding is directly added with the positional encoding as the final representation:
$$z_i=WE(x_i)+PE(i),$$ 
where $x_i$ is the token at the $i$-th position, $WE$ is the word embedding, and $PE$ is the positional encoding, which can be either a learnable embedding or a pre-defined function.

\paragraph{Multi-Head Self-Attention}
The attention mechanism is often used in an encoder-decoder architecture, and there are many variants of attention implementations \citep{bahdanau2014neural, britz2017massive}.
In Transformers, the \emph{scaled dot-product attention} is applied:
$$\text{attention}(Q,K,V)=\text{softmax}(\frac{QWK^TW}{\sqrt{d_k}})VW,$$
where $W$ is a linear projection and $Q$, $K$, $V$ represent query, key and value matrices respectively.

Transformer blocks are composed of multi-head self-attention. Literally, the inputs $Q$, $K$, $V$ are the same and the attention is performed multiple times, and then the output heads are concatenated as the final output hidden state $h$. This process can be formulated as 
$$\text{head}_i = \text{attention}(Q,K,V)$$
$$h = \text{concat}([\text{head}_1, ... ,\text{head}_n])W.$$

\paragraph{Transformer Encoder}
A Transformer encoder layer is composed of multi-head self-attention following a position-wise feed-forward network (FFN) with the residual connection \citep{he2016deep} and layer normalization~\cite{ba2016layer}:
$$\text{output}=\text{layernorm}(h + \text{FFN}(h)),$$
and then stacked the layers sequentially to form a Transformer encoder.

\paragraph{Transformer Decoder}
The Transformer decoder is also stacked by self-attention blocks, and it only has two major differences from the encoder:
\begin{enumerate}
    \item Each Transformer decoder layer has an additional sub-layer to perform attention on the encoder output.
    \item To ensure the decoder can only decode tokens depending on the tokens in the past, it uses an attention mask to mask the attention values of the subsequent tokens.
\end{enumerate}
Therefore, the Transformer decoder can decode tokens autoregressively like other conventional language models such as RNN.

\section{Position Embedding Analysis}
\label{sec:pe_analysis}
In this section, we conduct feature-level analyses of the pre-trained position embeddings of two Transformer encoders: BERT \citep{devlin2018bert} and RoBERTa \citep{liu2019roberta}, one Transformer decoder: GPT-2 \citep{radford2019language}, and also the sinusoidal function proposed by \citet{vaswani2017attention} is defined as
\begin{align*}
PE_{(i,2j)}&=\sin(i/10000^{2j/d_{model}}),\\
PE_{(i,2j+1)}&=\cos(i/10000^{2j/d_{model}}),
\end{align*}
where $i$ is the position index and $j$ is the dimension index.

\subsection{Do Embeddings Learn the Meaning of Positions?}
\label{sec:mean_of_pos}
Given the position space $\mathcal{P}$ and the embedding space $\mathcal{X}$, the goal of the position embedding function is to learn a mapping $f:\mathcal{P}\rightarrow\mathcal{X}$.
In the following experiments, we focus on answering two questions for better understanding what the embeddings capture:
\begin{enumerate}
    \item Can the learned embedding space $\mathcal{X}$ represent the absolute positions of the words?
    \item Are $\mathcal{P}$ and $\mathcal{X}$ isomorphic?
\end{enumerate}

\subsubsection{Absolute Position Regression}
If a position embedding can actually capture its \emph{absolute position}, 
it should be easy to reconstruct a reversed mapping function $g:\mathcal{X}\rightarrow\mathcal{P}$.
Thus, we use linear regression to learn a function $g$ that transfers the embeddings to the original positions. The feature dimension is $768$, and the maximum position in GPT-2 is trimmed from $1024$ to $512$ for comparison which BERT and RoBERTa.
Because we only have $512$ data points for each learned embedding, a 5-fold cross-validation is applied to avoid overfitting. 
The reversed mapping functions are evaluated by \textbf{Mean Absolute Error (MAE)}, and the result is shown in Table \ref{tab:abs_regression}.

From the results, the reversed mapping function of sinusoid can perfectly represent the absolute positions, and GPT-2 only has a small error. 
In contrast, the embeddings learned by Transformer encoders do not learn the information about the absolute positions, especially BERT which has an extremely high mean absolute error.

Additionally, we have also tried some more complicated non-linear models such as SVM or MLP to map the embeddings back. However, they easily overfit and the testing results are even worse than linear models. This implies that the position information in Transformer can actually be modeled by a linear model.

\begin{table}[t]
    \centering
    \begin{tabular}{c|lr}
        \toprule
        Type & PE & MAE \\
        \midrule
        \multirow{3}{*}{Learned} & BERT &  $34.14$\\
        & RoBERTa & $6.06$\\
        & GPT-2 & $1.03$\\
        \midrule
        Pre-Defined & sinusoid & $0.0$\\
        \bottomrule
    \end{tabular}
    \caption{Mean absolute error of the reversed mapping function learned by linear regression.
    }
    \label{tab:abs_regression}
    \vspace{-1mm}
\end{table}

\begin{table}[t]
    \centering
    \begin{tabular}{c|lr}
        \toprule
        Type & PE & Error Rate \\
        \midrule
        \multirow{3}{*}{Learned} & BERT &  $19.72\%$\\
        & RoBERTa & $7.23\%$\\
        & GPT-2 & $1.56\%$\\
        \midrule
        Pre-Defined & sinusoid & $5.08\%$\\
        \bottomrule
    \end{tabular}
    \caption{Error rate of the relative position regression. 
    }
    \label{tab:rel_regression}
    \vspace{-1mm}
\end{table}
\begin{figure*}[t!]
\centering
\includegraphics[width=1.0\textwidth]{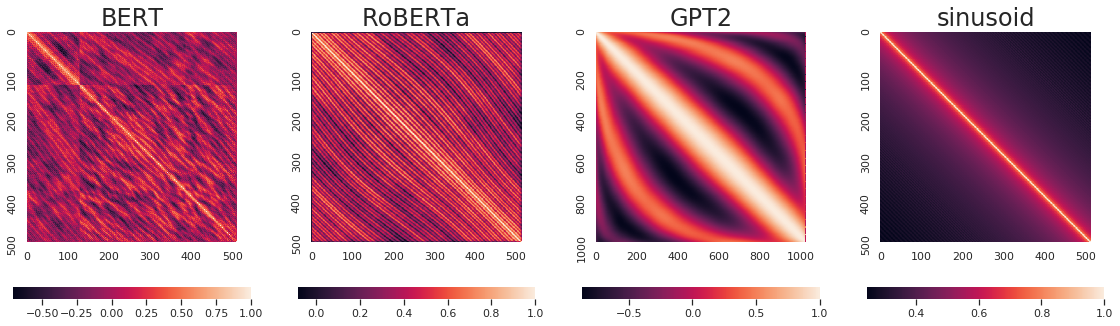}
\caption{Visualization of position-wise cosine similarity of different position embeddings.
Lighter in the figures denotes the higher similarity.
}
\label{fig:visualize}
\vspace{-2mm}
\end{figure*}

\subsubsection{Relative Position Regression}
In addition to absolute positions, the relation between positions is also informative (\emph{relative positions}). If $\mathcal{P}$ and $\mathcal{X}$ are isomorphic, there should exist a bijection of distance operation between two spaces. 
Thus we define a mapping function of distances from $\mathcal{X}$ to $\mathcal{P}$ : $h(x_i, x_j) = \|i - j\|$, where $i$, $j$ are two position indices, $\|i-j\|$ is the distance between $i$ and $j$ in the space $\mathcal{P}$, and $x_k$ is the position embedding at the $k$-th position.
In this scenario, we can also build a mapping function to check whether the embeddings can capture the relation between positions.
However, in our preliminary experiments, we find that using linear regression to predict the distance of positions is too hard, since the relation between positions in the space $\mathcal{X}$ may not be completely linear. 
Hence, we simplify this problem to \emph{whether the embeddings capture the order of every two positions}:
$$h(x_i, x_j) = \left\{\begin{array}{lr}
1 & \text{if } i \geq j\\
0 & \text{if } i < j
\end{array}\right\},$$
and then use logistic regression to learn this binary classification problem.

The results in Table \ref{tab:rel_regression} show that the position embeddings of Transformer encoders still learn less information about their position relations, especially for BERT.
Moreover, the sinusoid function, which can represent absolute positions perfectly, has a higher error rate than GPT-2 in relative positions but better than Transformer encoders, indicating the surprising capability of capturing such relations in GPT-2.

\subsection{What do Transformer Encoders Capture about Positions?}
According to the previous analyses, Transformer encoders (BERT and RoBERTa) may not well capture the meaning of positions (absolute and relative positions).
Therefore, the interested question becomes ``\emph{what do Transformer encoders capture about positions?}''.

\subsubsection{Position-Wise Cosine Similarity} 
\label{sec:visualize}
Figure \ref{fig:visualize} shows the visualization of position-wise cosine similarity of each position embedding. The point at $(i, j)$ indicates the similarity between the $i$-th position and the $j$-th position. 
First, we observe that the embedding space of sinusoid and GPT-2 have obvious periodic patterns along with position orders, which aligns the findings in the section \ref{sec:mean_of_pos}, where these two embeddings can actually capture the meanings of positions.
With regard to BERT, we can only observe that embedding vectors are similar to the positions nearby but have no explainable patterns in long-term relations.
Also, another observation is that BERT position embeddings have an obvious gap at the position $128$, because the pre-trained procedure of BERT trains on the sentences with the length of $128$ in the first stage, and then extends to the length of $512$ in the second stage.
The figure illustrates that the learned position information in the first stage can not be completely generalized to the second stage. Last but not least, the visualization of RoBERTa is similar to BERT, but have some limited non-periodic visible patterns at the positions nearby.

\begin{figure}[t!]
\centering
\includegraphics[width=0.5\textwidth]{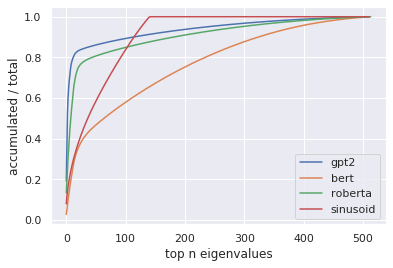}
\caption{Accumulated top eigenvalues of position embeddings. 
}
\label{fig:svd}
\vspace{-2mm}
\end{figure}
\subsubsection{Informativeness of Position Embeddings}
\label{sec:svd}
In order to examine the informativeness of the learned position embeddings, we apply singular value decomposition (SVD) on position embeddings and analyze their eigenvectors. 
Figure \ref{fig:svd} shows the curves of accumulated top $n$ eigenvalues versus the proportion of total eigenvalues.
Mathematically, the summation of the top $n$ eigenvalues indicates how informative can a matrix be if the matrix is transformed into a $n$-dim space. 
For a position space $\mathcal{P}$, which is a 1-dim space, we may not need a high-dimension embedding space $\mathcal{X}$ to represent the positions.
Thus, the summation of the top $n$ eigenvalues in a position embedding should account for the most proportion of total eigenvalues with only a very small $n$. 
However, in Figure \ref{fig:svd}, we find that the position embeddings of BERT take a very large $n$ to achieve a high proportion of total eigenvalues, and RoBERTa also takes a larger $n$ than GPT-2 and sinusoid. 
This implies that the position embeddings of Transformer encoders may learn more complex information rather than only about positions, and this rich information may only be useful in Transformer encoders.
This assumption will be further investigated in the experiments.

\subsection{What Make the Differences?}
Thus far, it can be found that the learned position embeddings between Transformer encoders and decoders are completely different. In this section, we will illustrate what makes these embeddings different.

\subsubsection{Pre-Training Objectives}
\label{sec:objective}
One of the main reason for the difference is the pre-training objectives.
Pre-trained Transformer encoders minimize the masked language modeling loss, which can be formulated as 
$$\mathcal{L}(\mathcal{U}) = \sum\limits_i \log P(u_i|u_1, ... u_{i-1}, u_{i+1}, ... u_l; \theta),$$
where $\mathcal{U}=\{u_1, ...u_l\}$ is the pre-training corpus with the length $l$, and $\theta$ is the model parameters. 
For Transformer decoders, the objective is the traditional autoregressive language modeling loss:
$$\mathcal{L}(\mathcal{U}) = \sum\limits_i \log P(u_i|u_1, ... u_{i-1}; \theta).$$

Transformer encoders can predict tokens depending on the tokens in both directions, while decoder can only predict depending on the token in the past. With enough context information, it is believed that Transformer encoders can succeed to predict tokens by only performing attention on the tokens nearby. That is why position embeddings learned by Transformer encoders do not need to involve the precise position information, aligning with the previous experiments in section~\ref{sec:mean_of_pos}.

We infer that encoder position embeddings may capture the local position information, which can force the output capturing the positions nearby, especially BERT almost involving nothing about absolute positions.
The inference makes the previous observations in sections \ref{sec:svd} and \ref{sec:visualize} sensible and explainable, and we will verify this inference through empirical experiments in section \ref{sec:nlp_tasks}.

\subsubsection{Differences Between BERT and RoBERTa}
\begin{figure}[t!]
\centering
\includegraphics[width=0.5\textwidth]{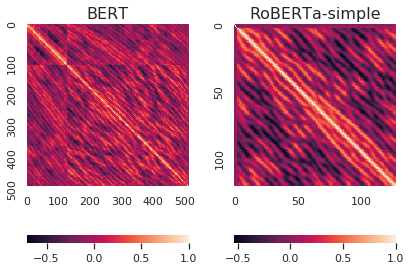}
\caption{Visualized position-wise cosine similarity of the simplified RoBERTa position embeddings.}
\label{fig:roberta_simple}
\vspace{-2mm}
\end{figure}
\label{sec:bert_roberta}
Both BERT and RoBERTa use the same Transformer encoder architecture, but there are still some obvious differences in the previous discussion.
Hence, we want to know what makes the position embeddings of BERT and RoBERTa different. 
The main improvements from BERT to RoBERTa are \cite{liu2019roberta}:
\begin{enumerate}
    \item Sequentially increasing the batch size from $256$ to $8192$ during training.
    \item Dynamically changing the token masks.
    \item Eliminating the additional next sentence prediction (NSP) loss.
\end{enumerate}

Due to the limitation of computing resources for experiments with a large batch size, we instead train a simplified version of RoBERTa that remains the $256$ batch size and the shorter block length for faster convergence. The visualized position-wise cosine similarity of the simplified RoBERTa position embeddings showing in Figure \ref{fig:roberta_simple} is very similar to BERT but without a gap at the position $128$. 
As a result, we infer that a large batch pre-training can make the position embedding more robust and involve more clear position information in Transformer encoders.

\begin{table*}[t!]
    \centering
    \begin{tabular}{lcc|cccccc}
        \toprule
        \bf PE & \bf Average & $\Delta$ & \bf SUBJ & \bf SST2 & \bf CR & \bf MR & \bf TREC & \bf MPQA \\
        \midrule
        Random & $0.7797$ & - & $0.8638$ & $0.7166$ & $0.7313$ & $0.7039$ & $0.8520$ & $0.8104$ \\
        \midrule
        BERT & $0.7868$ & $(+0.0071)$ & $0.8753$ & $0.7221$ & $0.7388$ & $0.7142$ & $0.8540$ & $0.8125$ \\
        RoBERTa & $0.7886$ & $(+0.0089)$& $0.8820$ & $0.7353$ & $0.7491$ & $0.7266$ & $0.8380$ & $0.8004$ \\
        GPT-2 & $0.7969$ & $(+0.0172)$ & $\textbf{0.8845}$ & $0.7446$ & $\textbf{0.7581}$ & $0.7314$ & $0.8540$ & $0.8087$ \\
        sinusoid & $\textbf{0.7983}$ & $\textbf{(+0.0186)}$ & $0.8801$ & $\textbf{0.7474}$ & $0.7549$ & $\textbf{0.7369}$ & $0.8580$ & $0.8125$ \\
        \midrule
        \multicolumn{3}{c|}{Average Length} & $23$ & $19$ & $19$ & $20$ & $10^\dagger$ & $3^\dagger$\\
        \bottomrule
    \end{tabular}
    \caption{Testing accuracy of text classification. 
    $\dagger$ indicates the much shorter average length in TREC and MPQA, so position embedding can not significantly affect the result.}
    \label{tab:nlu}
    \vspace{-2mm}
\end{table*}

\section{Performance Effect}
\label{sec:nlp_tasks}
In addition to the behavior analysis, we are interested in the performance difference of positional embeddings for different NLP tasks, where we conduct text classification (encoding), language modeling (decoding) and machine translation (encoding and decoding).
Note that each chosen task has its own important property where position information may cause different effects in Transformers.

\subsection{Text Classification}
Generally, for a text segment $s=\{x_1, x_2, ...x_n\}$ containing $n$ tokens, a Transformer for classification can be formulated as
\begin{align*}
h^0 &= \left[z_1, ..., z_n\right],\\
h^i &= \text{transformer\_block}(h^{i-1}), \\ 
P(y\mid s)&= \text{softmax}(h^l_n),
\end{align*}
where $z_i$ is the representation for the token $x_i$, $y$ is the output class and $h^i$ is the $i$-th layer output hidden state in Transformers. 

Conventionally, a special token, usually \texttt{[eos]} or \texttt{[CLS]} would be appended to the end of input tokens, so that the output hidden state can perform attention on all other input tokens.
In other words, no matter an encoder or a decoder is applied, the attention mask of the output hidden state and the objective can be identical. 
Therefore, we conduct a fair comparison with pre-trained position embeddings of both encoders and decoders in order to check whether all settings achieve similar performance.

\paragraph{Experimental Setup}
We experiment on six common text classification datasets: SST2, TREC, SUBJ, CR, MR, and MPQA. Since the last four datasets have no train/dev/test splits, we evaluate them with 5-fold cross-validation. We use the same model architecture as \citet{wang2019encoding}, building a 1 layer Transformer encoder with $256$ and $512$ hidden size for self-attention and feed-forward respectively and $8$ attention heads. 
Then five settings of the initialized position embeddings are performed: random, BERT, RoBERTa, GPT-2, and sinusoid, and other weights are initialized randomly.

\paragraph{Discussions}
Table \ref{tab:nlu} shows the results of text classification accuracy. BERT and RoBERTa position embeddings perform much worse than GPT-2 and sinusoid in most cases.
Because the output hidden state can utilize the information of all input tokens, the importance of absolute positions is certainly greater than local position information.
However, in TREC and MPQA, the difference between 5 settings is insignificant, and we notice that the average lengths of these two sets are much shorter than others shown in the bottom of Table \ref{tab:nlu}.
Therefore, the position information is not very important in these tasks (TREC and MPQA), considering that the local positions or even random initialization can result in the performance as well as one with absolute positions.
The experiments imply that even though text classification allows the model to utilize all tokens when making the prediction, the absolute positions, which GPT-2 can capture, may be still salient for longer inputs.

\paragraph{Length Sensitivity}

\begin{figure}[t!]
\centering
\includegraphics[width=0.48\textwidth]{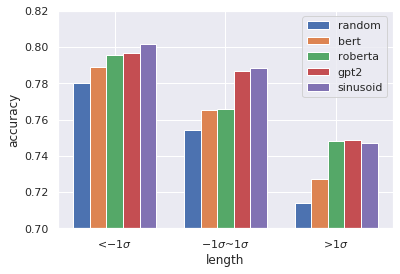}
\caption{Length versus accuracy in text classification.
}
\label{fig:nlu_length}
\vspace{-2mm}
\end{figure}
\begin{table*}[t!]
    \centering
    \begin{tabular}{c|l|cccc}
    \toprule
    \multirow{2}{*}{\bf LM} & \multicolumn{1}{c|}{\multirow{2}{*}{\bf PE}}& \multicolumn{2}{c}{\bf Wikitext-2} & \multicolumn{2}{c}{\bf Wikitext-103}\\
    & &\bf Perplexity & $\Delta$ & \bf Perplexity & $\Delta$ \\
    \midrule
    \multirow{4}{*}{MLM}&BERT & $147.93$ & - & $12.45$ & - \\
    &\quad+skip position & $198.61$ & $(+50.68)$ & $323.12$ & $(+310.67)$\\
    &RoBERTa & $157.98$ & - & $12.61$ & -\\
    &\quad+skip position & $199.13$ & $(+41.14)$ & $14.44$ & $(+1.83)$\\
    \midrule
    \multirow{2}{*}{Autoregressive}&GPT-2 & $172.97$ & - & $25.83$ & -\\
    &\quad+skip position & $171.20$ & $(-1.77)$ & $25.74$ & $(-0.09)$\\
    \bottomrule
    \end{tabular}
    \caption{Testing perplexity in Wikitext-2 and Wikitext-103. 
    }
    \label{tab:lm}
    \vspace{-2mm}
\end{table*}
To further analyze how the position embeddings affect text classification with different sentence lengths, we plot different ranges of lengths versus accuracy in Figure \ref{fig:nlu_length}.
Here we only calculate the average accuracy of SUBJ, SST, and CR since the average lengths of TREC and MPQA are too short. 
MR dataset is also excluded, because we find the distribution of length and accuracy in MR is too different from other three datasets and it may cause a huge bias in the figure.
Note that the results of MR roughly agrees with others.

In Figure \ref{fig:nlu_length}, sinusoid and GPT-2 still have higher accuracy with the length shorter than one standard deviation of the whole dataset, but the difference is very subtle.
In contrast, there is a significant gap between Transformer encoders and GPT-2 in longer sentences. In terms of extremely long sentences (longer than one standard deviation), we can only observe that BERT and random initialization perform much worse than others. We consider that the data distributions in this range have a too large bias so the results may not be robust.
Therefore, the analysis provides a hint that GPT-2 may be better to tackle the longer inputs for classification.

\subsection{Language Modeling}
In section \ref{sec:objective}, we have introduced the objectives of the masked language model and autoregressive language model for Transformer encoders and decoders respectively. Also, in the previous discussions, it is believed that the masked language model only learns the local position information to make the output tokens capture the positions nearby. To further verify this inference, we propose the \textbf{skip position attack} on position embeddings.

\paragraph{Skip Position Attack}
We propose a \emph{skip position attack} that skips the position index of input tokens. Originally, the input embedding can be represented as 
$$z_i=WE(x_i) + PE(i)$$.
However, in this attack, we multiply the input position index by a constant $k$, then the input embedding of token $x_i$ becomes
$$z_i=WE(x_i) + PE(i*k).$$
If the embedding only learns the local position information, skip position attack will skip the position indices nearby and lose local information. On the other hand, the absolute positions will not be influenced so much, because the order of the skipped positions is still the same.
Based on the design, we conduct experiments to validate our inference.

\paragraph{Experimental Setup}
We conduct the experiments on the Wikitext-2 and Wikitext-103 datasets, which have 2 million and 103 million tokens respectively. For model architecture, we take \texttt{BERT-Base} for a masked language model and \texttt{GPT-2-Base} as an autoregressive language model, both models have $12$ Transformer layers and $768$ hidden size. Similar to text classification, all weights are randomly initialized except position embeddings.
The constant $k$ in the skip position attack is set to $4$, and we slice the corpus into blocks of $128$ length to fit the maximum length of pre-trained position embeddings, which is $512$.

\paragraph{Discussions}
Table \ref{tab:lm} shows the results. On average, the masked language models (BERT and RoBERTa) have slightly lower perplexity than the autoregressive language model (GPT-2) due to their bidirectional token dependency. 
However, the skip position attack significantly harms the performance of the masked language models, while it affects nothing on the autoregressive language model.

Another observation is that, in Wikitext-2, the distribution of position information is not robust enough so the difference between position embeddings of BERT and RoBERTa is not significant.
However, in the larger dataset: Wikitext-103, skip position attack leads BERT position embeddings to extremely awful performance.
The observation here is consistent with the inferences mentioned in section \ref{sec:pe_analysis}, and we can conclude that position embeddings of Transformer encoders focus on capturing the information nearby, especially BERT, which involves even less position information than RoBERTa.

\subsection{Machine Translation}
Neural machine translation is often trained by a sequence-to-sequence model \citep{sutskever2014sequence}, which includes both encoder and decoder in the model. Thus, there are two position embeddings in a Transformer for machine translation, and the position embeddings in the encoder and in the decoder may cause different effects in this task.

\paragraph{Experimental Setup}

We experiment on the Multi30k English-German dataset from WMT2016 shared tasks. The properties of the dataset are shown in Table \ref{tab:muulti30k}. We use the scripts implemented by Fairseq \citep{ott2019fairseq} for a faster training process. The encoder and decoder have both $6$ layers where each layer has $4$ heads, $512$, and $1024$ hidden size for attention head and feed-forward respectively. Also, byte-pair encoding (BPE) \citep{sennrich2015neural,gage1994new} is applied to the corpus and the vocabulary size is reduced to $10,000$.

\begin{table}[h!]
    \centering
    \begin{tabular}{lccc}
        \toprule
         & \bf Train & \bf Valid & \bf Test \\
         \midrule
         Sentence Pairs & $29,000$ & $1,015$ & $1,000$ \\ 
         Average Length & $12$ & $12$ & $12$ \\
        \bottomrule
    \end{tabular}
    \caption{Statistics of Multi30k dataset.}
    \label{tab:muulti30k}
\end{table}

To respectively investigate the effectiveness on the encoder and the decoder, there are total four different initialization settings of pre-trained position embeddings:
\begin{compactenum}
    \item Position embeddings only for the encoder
    \item Position embeddings only for the decoder
    \item Different types of position embeddings for the encoder and decoder
    \item Same position embeddings for both encoder and decoder
\end{compactenum}
For the first three settings, only BERT and GPT-2 are performed for conciseness.

The results are shown in Table \ref{tab:translation}, where we evaluate the BLEU scores on the sentences longer than $2$ standard deviation (for both source and target) to analyze the effectiveness of longer sentences with consideration that the average length of Multi30k is relatively short.

\begin{table}[t!]
    \centering
    \begin{tabular}{ll|rr}
    \toprule
    
    \multicolumn{2}{c|}{\bf PE}& \multicolumn{2}{c}{\bf BLEU} \\
    \multirow{2}{*}{\bf Encoder} & \multirow{2}{*}{\bf Decoder} & \multirow{2}{*}{\bf Full Set} & \bf Length \\
    & & & $> 2\sigma$ \\
    \midrule
    Random & Random & $32.19$ & $18.04$\\
    \midrule
    BERT & \multicolumn{1}{c|}{-} & $35.54$ & $22.98$\\
    GPT-2 & \multicolumn{1}{c|}{-} & $34.36$ & $22.05$\\
    \midrule
    \multicolumn{1}{c}{-} & BERT & $32.08$ & $17.77$\\
    \multicolumn{1}{c}{-} & GPT-2 & $32.81$ & $18.05$\\
    \midrule
    BERT & GPT-2 & $34.11$ & $21.29$ \\
    GPT-2 & BERT & $32.80$ & $21.96$ \\
    \midrule
    BERT & BERT & $\textbf{35.94}$ & $23.60$\\
    RoBERTa & RoBERTa & $35.47$ & $24.50$ \\
    GPT-2 & GPT-2 & $35.80$ & $\textbf{25.12}$\\
    \bottomrule
    \end{tabular}
    \caption{BLEU scores on full set and long sentences ($>2\sigma$) of Multi30k translation data. The hyphen (-) in the table means the same as the baseline (random).
}
    \label{tab:translation}
\end{table}

\paragraph{Encoder}
Both BERT and GPT-2 position embedding can be effective in the encoder, especially BERT. The reason is that the decoded tokens can perform attention on all encoder outputs, thus the objective would be similar to the masked language modeling.

\paragraph{Decoder}
The effectiveness of position embeddings in the decoder is not as significant as one in the encoder, because the decoder cannot capture the order of the source language.
We also observe that applying BERT position embeddings on the decoder even slightly harms the performance, since it may make the decoder tend to focus on the tokens nearby only.

\paragraph{Different for Encoder/Decoder}
According to the previous results, we hypothesize that using BERT in the encoder and GPT-2 in the decoder could perform best. 
However, in our experiments, using different pre-trained position embeddings is even worse, probably because the divergence of position embeddings trained by different models is quite huge and mixing them in the same model may not suitable.
Also, we swap the position embeddings in the encoder and decoder to see the impact. The BLEU score of the full set drops a lot, but in terms of long sentences, using GPT-2 in the encoder may not lose too much performance.

\paragraph{Same for Encoder/Decoder}
The results show that the performance between three pre-trained position embeddings are very close in the full set.
However, in terms of longer sentences, GPT-2 is much better than BERT and RoBERTa. This observation aligns well with the previous analysis that the absolute position information is more important for longer sentences.

To sum up, there main observations are found: 1) The effectiveness of position embeddings in the encoder is more significant than one in the decoder. 2) Mixing different position embeddings in a model is not suitable. 3) GPT-2 position embeddings outperform others when modeling longer sentences.

\section{Conclusion}
This paper investigates the implicit meaning of pre-trained Transformer position embeddings. Transformer encoders learn the local position information that can only be effective in masked language modeling. On the other hand, the Transformer decoders for autoregressive language modeling actually learn about absolute positions. 
The empirical experiments on the pre-trained position embeddings validate our hypothesis.
We also show that different NLP tasks with different model architectures and different training objectives may utilize the position information in different ways.
As a result, it is believed that this study will benefit future work about choosing suitable positional encoding functions or designing other modeling methods for position information in the target NLP tasks based on their properties.

\section*{Acknowledgements}
We thank reviewers for their insightful comments.
This work was financially supported from the Young Scholar Fellowship Program by Ministry of Science and Technology (MOST) in Taiwan, under Grant 109-2636-E-002-026.

\bibliographystyle{acl_natbib}
\bibliography{anthology,emnlp2020}
\newpage
\appendix
\section{Reproducibility}
\subsection{Datasets}
The datasets we used can be downloaded from the following linked pages, and the details of datasets are also described in the pages.
\begin{itemize}
    \item Text Classification:
    \href{https://reurl.cc/yZXKdO}{link}
    \item Language Modeling: 
    \href{https://reurl.cc/L3DQyy}{link}
    \item Machine Translation: 
    \href{https://reurl.cc/exaVex}{link}
\end{itemize}
\subsection{Training Details}
\begin{table}[h]
    \centering
    \textbf{Text Classification}
    \vspace{2mm}
    
    \begin{tabular}{c|c}
        \toprule
         tokenizer & spacy \\
         optimizer & Adam \\
         lr & $1^{-4}$ \\
         batch\_size & $32$ \\
         max\_epoch & 40 \\
         \bottomrule
    \end{tabular}
\end{table}
\begin{table}[h]
    \centering
    \textbf{Language Modeling}
    \vspace{2mm}
    
    \begin{tabular}{c|c|c}
        \toprule
         & Wikitext02 & Wikitext-103\\
         \midrule
         lr & - & $2.5^{-4}$ \\
         batch\_size & $32$ & $32$ \\
         max\_epoch & $20$ & $3$ \\
         warup\_steps & - & $4000$ \\
         \bottomrule
    \end{tabular}
\end{table}

\begin{table}[h]
    \centering
    \textbf{Machine Translation}
    \vspace{2mm}
    
    \begin{tabular}{c|c}
        \toprule
         optimizer & Adam \\
         weight\_decay & 0.0001 \\
         lr & $1^{-4}$ \\
         max\_tokens & $2048$ \\
         max\_epoch & $40$ \\
         lr\_scheduler & inverse\_sqrt \\
         warup\_steps & $4000$ \\
         label\_smooth & $0.1$ \\
         \bottomrule
    \end{tabular}
\end{table}
Since the goal of this paper is to compare position embedding, we do not try too many hyperparameters on Transformers, and most setting are default as the implementation of \href{https://github.com/huggingface/transformers/tree/master/examples/language-modeling}{hugginface} and \href{https://github.com/pytorch/fairseq}{Fairseq}.

\subsection{Running Time}
All our experiments are trained on 1 GTX 2080 TI GPU. Except language modeling on Wikitext-103 takes about 10 hours, all other trainings can be done within 2 hours.

\end{document}